\documentclass[letterpaper, 10pt, conference]{ieeeconf}  

\IEEEoverridecommandlockouts                              

\usepackage{graphics} 
\usepackage{graphicx}

\makeatletter
\let\MYcaption\@makecaption
\makeatother
\usepackage{subcaption}
\makeatletter
\let\@makecaption\MYcaption
\makeatother

\usepackage{xcolor}
\usepackage{lipsum}

\usepackage[T1]{fontenc}
\usepackage{siunitx}
\DeclareSIUnit{\sqrthertz}{\sqrt{\unit{\hertz}}}

\usepackage{booktabs}
\usepackage{multirow}
\usepackage{makecell}
\usepackage{threeparttable}

\usepackage[nolist,nohyperlinks]{acronym}
\acrodef{imu}[IMU]{Inertial Measurement Unit}
\acrodef{fov}[FoV]{Field of View}
\acrodef{tof}[ToF]{Time of Flight}
\acrodef{vtol}[VTOL]{Vertical Take-Off and Landing}
\acrodef{ros}[ROS]{Robot Operating System}
\acrodef{lan}[LAN]{Local Area Network}
\acrodef{ptp}[PTP]{Precision Time Protocol}
\acrodef{ntnu}[NTNU]{Norwegian University of Science and Technology}
\acrodef{oasis}[OASIS]{Omnidirectional Autonomy Sensor Integration System}
\acrodef{ve}[VE]{Volumetric Exploration}
\acrodef{gvi}[GVI]{General Visual Inspection}
\acrodef{ei}[EI]{Exploration and Inspection}
\acrodef{tsp}[TSP]{Traveling Salesman Problem}
\acrodef{uart}[UART]{Universal Asynchronous Receiver/Transmitter}
\acrodef{swap}[SWaP]{Size, Weight and Power}
\acrodef{slam}[SLAM]{Simultaneous Localization and Mapping}
\acrodef{vtol}[VToL]{Vertical Takeoff and Landing}
\acrodef{dof}[DoF]{Degrees of Freedom}
\acrodef{fmcw}[FMCW]{Frequency Modulated Continuous Wave}
\acrodef{ugv}[UGV]{Unmanned Ground Vehicles}
\acrodef{uav}[UAV]{Unmanned Aerial Vehicles}
\acrodef{sota}[SOTA]{State-of-the-Art}
\acrodef{mipi}[MIPI]{Mobile Industry Processor Interface}
\acrodef{csi2}[CSI-2]{Camera Serial Interface 2}
\acrodef{rmse}[RMSE]{Root Mean Square Error}
\acrodef{ape}[APE]{Absolute Pose Error}
\acrodef{rpe}[RPE]{Relative Pose Error}

\usepackage[hidelinks]{hyperref}
\usepackage[capitalise]{cleveref}

\usepackage{flushend}

\newcommand{\moduleName}{UniPilot}

\title{\LARGE \bf
\moduleName{}: Enabling GPS-Denied Autonomy Across Embodiments
}

\author{Mihir Kulkarni, Mihir Dharmadhikari, Nikhil Khedekar, Morten Nissov, Mohit Singh, \\ Philipp Weiss, and Kostas Alexis
\thanks{This work was supported by (a) AFOSR Award No. FA8655-21-1-7033, (b) the Horizon Europe grant SPEAR (101119774), and (c) the Horizon Europe grant AUTOASSESS (101120732), and }
\thanks{All authors are with the Department of Engineering Cybernetics, O. S. Bragstads Plass 2D, Norwegian University of Science and Technology (NTNU), Trondheim, Norway. {\tt\small mihir.kulkarni@ntnu.no}}
}

\newcommand{\insertfig}{
\includegraphics[width=\linewidth]{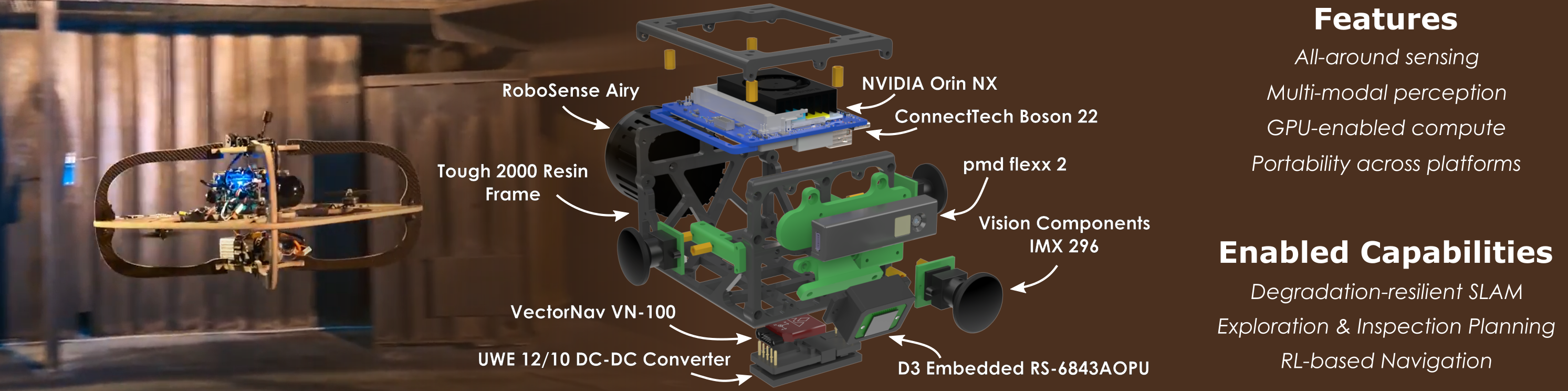}
\vspace{-2em}
\captionof{figure}{Exploded view of the \moduleName{} module, with hardware, sensing, and compute components highlighted, alongside an image of the aerial platform that integrates the module in a process tank environment.}
\label{fig:exploded_view}%
\vspace{-1em}
}

\makeatletter
\apptocmd{\@maketitle}{\centering\insertfig}{}{}
\makeatother

\begin{document}
\maketitle

\begin{abstract}
This paper presents \moduleName{}, a compact hardware-software autonomy payload that can be integrated across diverse robot embodiments to enable autonomous operation in GPS-denied environments. The system integrates a multi-modal sensing suite including LiDAR, radar, vision, and inertial sensing for robust operation in conditions where uni-modal approaches may fail. \moduleName{} runs a complete autonomy software comprising multi-modal perception, exploration and inspection path planning, and learning-based navigation policies. The payload provides robust localization, mapping, planning, and safety and control capabilities in a single unit that can be deployed across a wide range of platforms. A large number of experiments are conducted across diverse environments and on a variety of robot platforms to validate the mapping, planning, and safe navigation capabilities enabled by the payload.

\end{abstract}

\section{INTRODUCTION}
With the progressing rise of autonomy, it becomes increasingly more important to be able to test a wide variety of complex methods and systems on easily re-creatable platforms. Especially of interest is to investigate and demonstrate a method's generality across different robot embodiments. Furthermore, the demand for autonomy in different robot actualization and in varied environments necessitates flexibility at the core of this design philosophy. This has become clear with the rise of multi-system solutions considered for difficult challenges in autonomy \cite{tranzatto2022teamcerberuswinsdarpa,ebadi2024PastPresentFuture}.
As such, these robots need to integrate an appropriate sensing suite that facilitates perception in the environments of interest and a computing system capable of running the autonomy software. Designing such a payload requires careful consideration of sensor selection and placement, computing requirements, power limitations, and weight restrictions.

Motivated by the above, we present a plug-and-play autonomy payload called \moduleName{}. \moduleName{} is a hardware-software payload that can be integrated on aerial or ground robots for autonomous operations. The module is designed with autonomy at its core. \moduleName{} carries a multi-modal sensing suite integrating visual cameras, LiDAR, radar, \ac{tof} sensor, and an \ac{imu}, as well as an NVIDIA Jetson Orin NX compute board that runs the autonomy software comprising the multi-modal \ac{slam} solution, exploration and inspection path planning, and learned navigation policies. The compute and sensing capabilities allow users to test a wide range of methods while the compact and lightweight design of \moduleName{} enables deployment across a variety of aerial and ground robots.

\moduleName{} is extensively tested on a collision-tolerant aerial platform, a legged robot, and a hybrid Vertical Take-Off and Landing (VTOL) platform across a diverse set of environments. Specifically, we demonstrate the performance of the \ac{slam} solution in a high-speed (up to $\SI{14}{\meter \per \second}$) scenario and the use of multi-modality using a collision-tolerant quadrotor. Furthermore, the \ac{slam} performance is evaluated on the hybrid VTOL platform. Additionally, a mission demonstrating the learning-based navigation policy for safe navigation is presented. Finally, autonomous exploration and inspection missions are performed in a natural mine and two industrial environments with the collision-tolerant aerial robot, and autonomous exploration of an urban underground environment is achieved using a legged robot, demonstrating the autonomy readiness of \moduleName{}.

The remainder of the paper is organized as follows: Related works are presented in Section~\ref{sec:related_work}. Section~\ref{sec:system} describes the system design, including the hardware design and the components of the autonomy software. Experiments and deployments of \moduleName{} onboard various robots are presented in~\cref{sec:eval}. Finally, conclusions are drawn in Section~\ref{sec:conclusion}.

\section{RELATED WORK}\label{sec:related_work}



With the increased interest in the deployment of autonomous robots for different missions, there has been an increased need for sensing and autonomy payloads in research and industry~\cite{frey2025boxidesigndecisionscontext,2023leica}. These payloads provide varying capabilities ranging from pure sensing~\cite{2024everysync}, integrated \ac{slam} solution~\cite{2014_fpga_stereo}, to providing full autonomy capabilities~\cite{hovermap}.

Early works in this domain include the development of plug-and-play synchronized stereo cameras, and \ac{imu} systems that provided out-of-the-box visual \ac{slam} capabilities~\cite{2014_fpga_stereo,2018pirvs}. As perception is being tested in incrementally more difficult environments, the necessity for multi-modal sensing becomes evident \cite{ebadi2024PastPresentFuture}. As such, the payloads follow suit and integrate a wider array of sensing modalities. \cite{2024everysync} presents the open-hardware time synchronization unit, EverySync, along with a LiDAR, \ac{imu}, and camera payload synchronized with EverySync. \cite{frey2025boxidesigndecisionscontext} presents a sensing payload consisting of multiple LiDARs, stereo cameras, \ac{imu}s, and details the design procedure behind building such a payload. The payload CatPack is a perception payload consisting of RGB and thermal cameras, and a rotating LiDAR for \ac{ugv}.
Leica Geosystems BLK2Go is a commercial handheld mapping unit that integrates 3 cameras, an \ac{imu}, and an \qty{830}{\nano\meter} laser scanner with \qty{3}{\milli\meter} accuracy.

Going beyond purely sensing or \ac{slam} payloads, a set of solutions also provide autonomy capabilities. Hovermap~\cite{hovermap} presents a rotating LiDAR-based mapping payload for handheld or \ac{uav} mounted operations. Hovermap provides collision-avoidance functionality along with waypoint following with it's integrated LiDAR \ac{slam}. Similarly, Alphasense Autonomy~\cite{alphasense_autonomy} presents a synchronized multi-camera, \ac{imu} system integrating 3D visual \ac{slam}, collision-avoidance, and waypoint navigation capabilities. 

The majority of the current works present systems that serve as either a sensing unit -- with some providing integrated \ac{slam} capabilities -- or with a limited level of autonomy (collision-avoidance or waypoint navigation). Despite the maturity of research in autonomous exploration, inspection, and generally GPS-denied navigation, there is a lack of a unified solution in the form of a plug-and-play module that can provide these capabilities integrated with the sensing and computing unit, while still being an appropriate size and weight such as to facilitate usage onboard small aerial platforms. \moduleName{} bridges this gap by integrating \ac{sota} autonomy solutions into a multi-modal sensing and computing unit.

\section{SYSTEM DESCRIPTION}\label{sec:system}
The design of \moduleName{} is driven by specific requirements for multi-platform autonomy in GPS-denied environments. Key design constraints include: (1) total weight \SI{<1}{\kilo\gram} for compatibility with small aerial platforms, (2) power consumption \SI{<50}{\watt} for \SI{30}{\minute} missions, (3) operating temperature range \SI{-20}{\celsius} to \SI{60}{\celsius}, (4) multi-modal sensing with overlapping fields of view, and (5) plug-and-play integration via standard interfaces.

\subsection{Component Selection}\label{sec:component_selection}
The component selection for the module is motivated by challenges presented to autonomy across a diverse set of conditions ranging from large-scale environments, self-similar geometries, lack of visual texture to those presenting obscurants. A multi-modal sensing suite is employed to enable operation in such diverse environments. The sensors are placed to (1) provide all around perception and (2) ensure overlap between different sensing modalities as shown in~\cref{fig:fov-image}. Moreover, accurate surface mapping is considered key for inspection in industrial settings alongside purely visual inspection. To run most modern \ac{slam}, path planning, and control algorithms the module must also host sufficient compute. To enable use on aerial platforms, the weight limit of the module is restricted to \SI{1}{\kilo\gram}. Correspondingly, selection of sensors must satisfy the weight requirements and compatibility with the interfaces available on the compute carrier board.

\subsubsection{Compute}\label{sec:compute}
\begin{figure}[ht!]
    \centering
    \includegraphics[width=\linewidth]{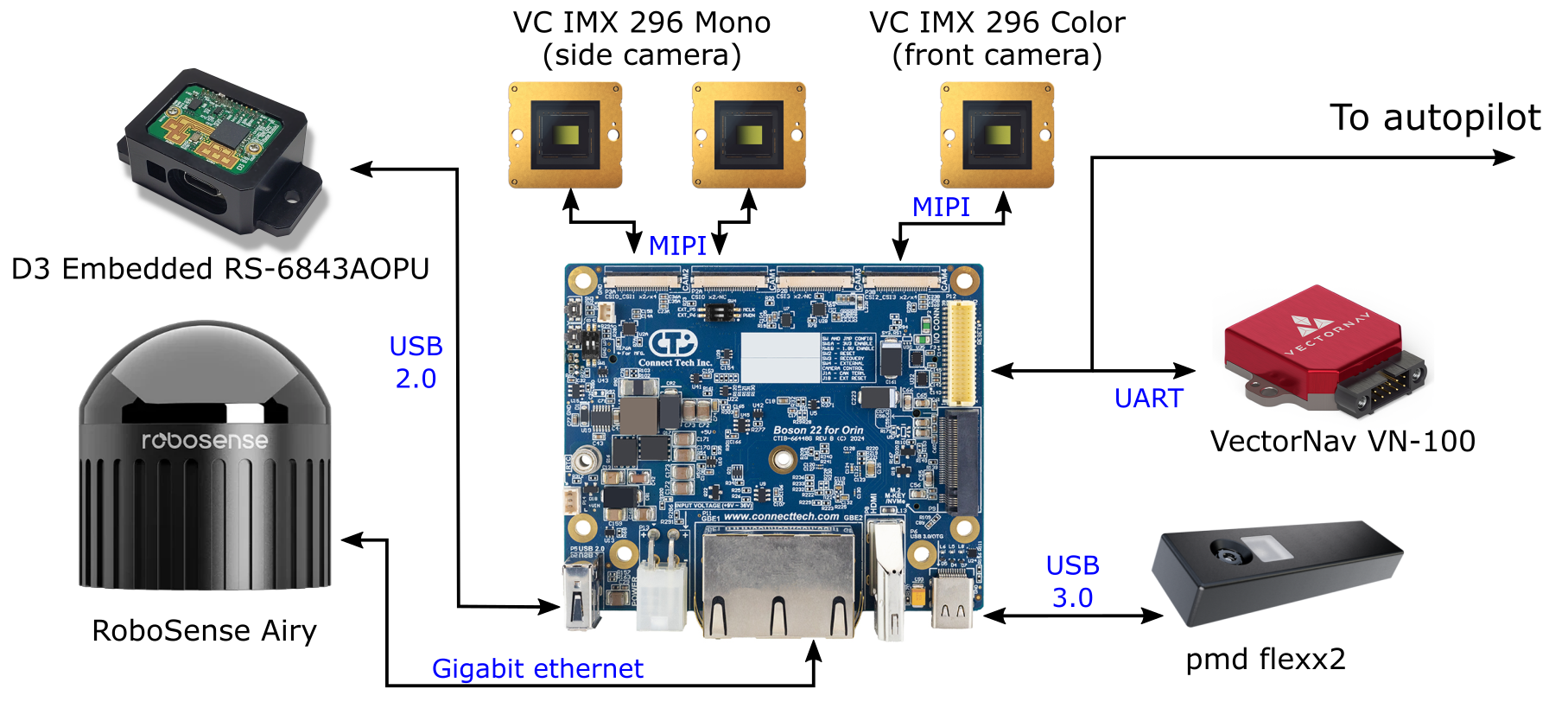}
    \caption{Visualization of the interfaces between different sensor modules (and autopilot) and the carrier board for the onboard computer.}
    \label{fig:connection-diagram}
\end{figure}

An NVIDIA Jetson Orin NX \SI{16}{\giga\byte} module is selected as the compute since it hosts a powerful 8-core Arm Cortex CPU offering the capability to run CPU-heavy software such as perception, mapping and planning algorithms alongside sensor drivers and a GPU offering up to $157$ TOPS of performance for performing onboard real-time inference for large neural networks. A ConnectTech Boson-22 carrier board~\cite{connecttechBoson22Carrier} was selected to mount the compute module due to the rich interfacing it provides (Table~\ref{tab:hardware}) while maintaining low \ac{swap} requirements. The carrier board supports four 22-pin MIPI CSI-2 connectors alongside a variety of other communication interfaces (e.g., USB, Ethernet, UART, CAN, I2C, SPI, etc). The interfaces between the sensors and the compute board is shown in~\cref{fig:connection-diagram}. The power to the module and the sensors is provided via a UWE 12/10 DC-DC Converter capable of providing up to \SI{120}{\watt} of power at \SI{12}{\volt}.

\subsubsection{Cameras}\label{sec:cameras}
To minimize the weight of the module, board-level camera sensors using the MIPI CSI-2 interface were chosen. Vision Components IMX-296 MIPI camera boards are used as they interface directly with the selected carrier board. Global shutter sensors were chosen over rolling shutter alternatives to prevent motion artifacts at high speeds. Rolling shutter sensors introduce distortion that are proportional to the product of the speed and the readout time, which are significant at our desired speeds (\SI{14}{\meter\per\second}). Two monochrome sensors are placed on the sides with large \ac{fov} lenses for all-around perception, while the color sensor is placed at the front of the module for inspection, colorization of mapped surfaces, and providing a live feed to the robot operator (if needed). Additionally for low-light environments, monochrome sensors can be binned to lower resolution for better light-sensitivity. 

\subsubsection{\acl{imu}}\label{sec:imu}
The VectorNav VN-100 \ac{imu} is interfaced with the compute module using the UART protocol. This sensor is configured to output sensor synchronization signals to allow precise correlation with the IMU samples. The sensor provides inertial measurements at a rate of \qty{200}{\hertz}. This sensor is chosen owing to its low-noise, stable bias characteristics, and the ability to output synchronization signals at a configurable rate.

\subsubsection{LiDAR}\label{sec:lidar}
RoboSense Airy is chosen as a low-cost, lightweight LiDAR solution, offering a $96$-beam sensor with a $90\unit{\degree}\times 360\unit{\degree}$ \ac{fov}, with a range of \qty{60}{\meter}. This sensor is interfaced with the compute platform using Gigabit Ethernet and synchronized with the compute module using IEEE 1588 Precision Time Protocol (PTP), and operated at a frequency of \SI{10}{\hertz}. The selection was based on systematic evaluation of weight (\SI{120}{\gram}), power (\SI{8}{\watt}), and vertical FoV requirements. Alternative solutions like Velodyne VLP-16 (\SI{830}{\gram}) exceed our weight budget, while solid-state options like Livox Mid-360 provides insufficient vertical coverage (\SI{59}{\degree}) for obstacle detection in confined spaces.

\subsubsection{Radar}\label{sec:radar}
A D3 Embedded RS-6843AOPU \ac{fmcw} radar is chosen based on its size, weight, \ac{fov} and angular resolution, which in contrast to typical automotive radars has larger elevation \ac{fov} and equal azimuth/elevation angular resolution. This sensor is interfaced with the computer using a USB 2.0 interface and is externally triggered by the VectorNav VN-100 IMU to maintain tight time synchronization.

\subsubsection{\acl{tof} Sensor}\label{sec:tof}
To accurately map inspected surfaces at a high resolution, a pmd flexx2 \ac{tof} camera is chosen and placed alongside the front-facing camera, owing to the accuracy of range measurements at a short range. Despite a low sensing range of \SI{4}{\meter}, the flexx2 sensor is able to perceive obstacles with narrow cross sections such as thin branches or wires, enabling its use for safe navigation. This sensor is connected to the computer using USB 3.0 and offers a high frame-rate (up to \SI{60}{\hertz}).

\subsection{Hardware Design}\label{sec:hardware_design}

\subsubsection{Component Placement}\label{sec:component_placement}
\begin{figure}[ht!]
    \centering
    \includegraphics[width=\linewidth]{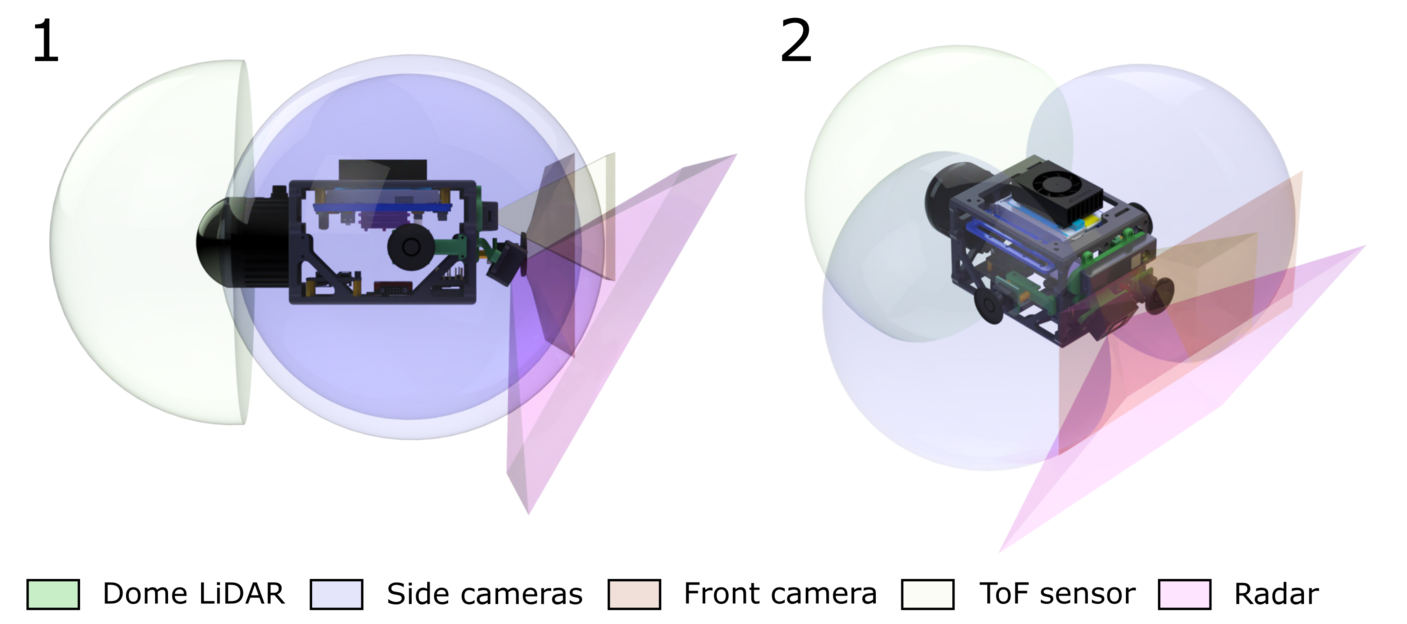}
    \caption{Visualization of the sensor FoV from dual-perspectives: (1) Side-view highlighting the FoV of the LiDAR, front-camera, ToF module, and the radar and (2) Oblique-view showing the FoV of all sensors on the module.}
    \label{fig:fov-image}
    \vspace{-3ex}
\end{figure}

The \moduleName{} is designed to be compatible with a wide range of robot platforms that include both aerial and legged robots. At the same time, it can also be operated as a hand-held sensor unit. To enable all-around sensing, the two monochrome cameras with the Vision Components IMX 296 sensors with high \ac{fov} fisheye lenses are placed on either sides of the module. These lenses provide a \SI{185}{\degree} \ac{fov}, and the placement of the cameras ensures that all regions around the module are captured. Considering the need for performing visual inspection and simultaneous mapping and reconstruction of surfaces, a Vision Components IMX 296 color camera with a wide-angle lens is placed at the front of the module, along with a pmd flexx2 ToF camera to map the visually inspected surfaces at a high resolution. The radar is mounted facing downward at an angle of \SI{30}{\degree} from the horizontal plane. This placement is chosen to ensure ground visibility for accurate velocity estimation for robust odometry. The RoboSense Airy dome-LiDAR sensor is placed at the back of the module to ensure that the features in the environment remain within the \ac{fov} of the LiDAR while the front of a robot is facing surfaces at close distances for inspection. Finally, the VectorNav VN-100 IMU is placed at the center of the base of the module to reduce the effects of vibrations. An exploded view of the sensors in the module can be seen in~\cref{fig:exploded_view}, while~\cref{fig:fov-image} shows the \ac{fov} of the sensors in \moduleName{}.

\subsubsection{Mechanical Design}\label{sec:mechanical_design}
In order to be compatible for mounting with a variety of platforms, \moduleName{} is designed with a standard set of $4$ screw holes on the bottom of the module. Designing platform-independent connections allows it to be mounted on a wide range of robots and handheld units. The module is manufactured using a Formlabs Form 3L stereolithography printer using the Tough 2000 resin material. The material is selected for its strength, impact resistance, thermal stability, electrical insulation properties and low weight, offering a good balance between mechanical robustness and weight. The total weight of \moduleName{} including the sensors, computer, power distribution, cabling, and structural material is \SI{829}{\gram}.

The bottom of the module is designed as three perpendicular plates with a top cover. The exteroceptive sensors are mounted on the vertical plates, while the IMU is mounted on the horizontal plate with the DC-DC converter. The vertical walls are also supported by ribs to increase strength and to reduce deformation with loading. The top cover is easily removable, allowing access to the internal components for maintenance and upgrades. 3D-printed mounts for the MIPI cameras are attached to the bottom of the module and placed outwards to ensure that the cameras are not obstructed by the module body or the other components on the module. The top of the module is a single plate that is only used to mount the compute unit and allows easy removal for maintenance.

\begin{table*}[ht]
    \centering
    \caption{Components Mounted on the \moduleName{} Module}
    \label{tab:hardware}
    \newcommand{\rotation}{0}
\newcommand{\spacing}{\ }
\begin{threeparttable}
\begin{tabular}{lllll}
    \toprule
        &Name   &{Weight [\unit{\gram}]}  &{Power [\unit{\watt}]}    &Description\\
    \midrule
    \multirow{1}{*}{\rotatebox{\rotation}{\textbf{Compute}}}
        &1x Orin NX   & 28 & 25    &{RAM: \SI{16}{\giga\byte}},\spacing{}{8-core Arm Cortex CPU},\spacing{}{1024-core NVIDIA GPU}\\
        &1x Boson-22 carrier board + heat sink   & 152 & 13.8    &\makecell[l]{4x \acs{mipi} \acs{csi2}, 2x \unit{\giga\bit} Ethernet, 1x USB 2.0, 1x USB 3.1,\\ 8x GPIO, 1x CAN, 2x I2C, 3x UART, 2x SPI interfaces}\\
    \midrule
    \multirow{1}{*}{\rotatebox{\rotation}{\textbf{\acs{imu}}}}
        &1x VectorNav VN-100   & 15 &  0.22  &{\qty{800}{\hertz}},\spacing{}{Accel.: \SI{140}{\micro\g\per\sqrthertz}},\spacing{}{Gyro.: \SI{0.0035}{\degree\per\second\per\sqrthertz}}\\
    \midrule
    \multirow{2}{*}{\rotatebox{\rotation}{\textbf{Cameras}}}
        &2x VC MIPI IMX296M   &  4 &  0.759  &{Grayscale},\spacing{}{Resolution: $1440\times1080$},\spacing{}{\acs{fov}: $185\unit{\degree}$}\\
        &1x VC MIPI IMX296C   &  4 & 0.759   &{Color},\spacing{}{Resolution: $1440\times1080$},\spacing{}{\acs{fov}: $118\unit{\degree}\times94\unit{\degree}$}\\
        &1x PMD Flexx2 \acs{tof}   & 13 &  0.68  &{Range: \qty{4}{\meter}},\spacing{}{\acs{fov}: $56 \times 44\unit{\degree}$}\\
    \midrule
    \multirow{1}{*}{\rotatebox{\rotation}{\textbf{Radar}}}
        &1x D3 Embedded RS-6843AOPU   & 25 & 7.5  &{Range: \qty{49}{\meter}},\spacing{}{\acs{fov}: $180 \times 180\unit{\degree}$}\\
    \midrule
    \multirow{1}{*}{\rotatebox{\rotation}{\textbf{LiDAR}}}
        &1x RoboSense Airy   & 230 &  8  &{Range: \qty{30}{\meter}},\spacing{}{\acs{fov}: $360 \times 90\unit{\degree}$}\\
    \bottomrule
\end{tabular}
\vspace{-2ex}
\end{threeparttable}

\end{table*}

\subsection{Time Synchronization}\label{sec:time_synchronization}
\begin{figure}
    \centering
    \includegraphics[width=1\linewidth]{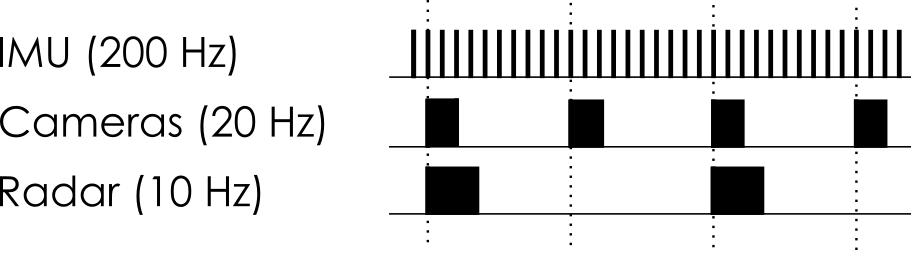}
    \caption{Visualization of the sampling of visual cameras and radar triggered by the \ac{imu}}
\label{fig:trigger_visualization}
\end{figure}

\moduleName{} utilizes different methods for time-synchronization, adapted to the individual sensor's particularities, to ensure that all measurements are timestamped accurately. The module uses the IEEE 1588 Precision Time Protocol (PTP) for time synchronization across Ethernet-based sensors (e.g., LiDARs) and the onboard computer. Furthermore, it exploits the capability of the VectorNav VN-100 \ac{imu} to generate trigger pulses at a configurable rate. This signal from the IMU is used to trigger sensors which expose a hardware pin for synchronization, such as the MIPI cameras and the radar as visualized in~\cref{fig:trigger_visualization}. The IMU measurements themselves are host-timestamped by the onboard computer, as there is little latency in sampling from the \ac{imu}. A buffer of timestamps corresponding to the synchronization signals is stored. The MIPI cameras and radar sensor drivers thus subscribe to the \ac{imu} trigger timestamps and solve the assignment problem using knowledge of the exposure and chirp durations as well as an a-priori calibration for approximate processing delays. The \ac{tof} sensor establishes its own timestamping through its driver.

\subsection{Autonomy Architecture}

\subsubsection{Perception}
The \moduleName{} module carries LiDAR, radar, vision, and \ac{imu} sensors, with the intent of enabling multi-modal estimation well-suited for robust performance across environments, despite perceptual degradations. This set was selected as a balance between the weight and power consumption of additional sensors and the complementary nature of the different modalities. For example, radar and LiDAR have a complementary relationship, as LiDAR can contribute accuracy and mapping density whereas radar can contribute robustness through lack of sensitivity to environment geometry, following \cite{nissov2024degradation,nissov2024Robust}.
For the experiments in question, the LiDAR and \ac{imu} are fused for simplicity using the geometry-only partition of \cite{khedekar2025pgliophotometricgeometricfusionrobust}. The output of the low-rate (\SI{10}{\hertz}) factor-graph optimization (the 6-DoF pose) is passed as a measurement to a general, high-rate fusion algorithm \cite{2013msf} to obtain an \ac{imu}-rate odometry for use by the onboard control.

\subsubsection{Planning}\label{sec:high_level_planning}
\moduleName{} integrates the exploration and inspection path planning method presented in~\cite{2023expgvi}, which is built on top of the open-sourced graph-based exploration planner (GBPlanner)~\cite{gbplanner2,GBPLANNER_JFR_2020}. 
Utilizing a volumetric map representation~\cite{oleynikova2017voxblox}, the planner presents two planning modes, namely \ac{ve} and \ac{gvi}. In the \ac{ve} mode, the planner starts with no prior map of the environment and aims to generate a 3D map of the given volume using a depth sensor. It operates in a bifurcated local-global planning architecture. 
The local planner calculates fast, efficient paths to maximize mapping of the unknown volume within a local area around the robot, while the global planner enables repositioning the robot to the frontiers of exploration as well as a safe return to the mission start point.

The \ac{gvi} mode enables inspection of the mapped surfaces of a given environment using a camera sensor at the desired viewing distance. It utilizes a volumetric map of the environment, obtained through the \ac{ve} mode or from a prior mission, to calculate the inspection path. A set of viewpoints is sampled at the desired viewing distance from the occupied surface of the volumetric map and connected with collision-free edges to form a 3D collision-free graph. Next, the minimal set of viewpoints providing complete coverage is selected, and the sequence to visit these viewpoints is calculated by solving the \ac{tsp}.

\subsubsection{Safe Navigation}\label{sec:safe_navigation}
Maintaining safe operations in spite of challenges in planning or estimation, such as insufficient resolution for small objects or inaccuracy due to degraded conditions, is critical for autonomous agents. As such, it is important that the \moduleName{} also demonstrates this capability. During operation in large-scale environments, the perception and localization algorithms are prone to drift. As a result, the volumetric map information used by the planner may be inconsistent, causing the paths being potentially planned through obstacles. An additional layer of safety is required to mitigate this challenge. In response to this, the \moduleName{} is equipped with a reinforcement learning-based safe navigation policy~\cite{kulkarni2024reinforcement} that commands velocity setpoints to the low-level controller. The policy uses partial robot state, a unit vector to goal location, and the raw depth images from the pmd flexx2 \ac{tof} sensor. The NVIDIA Orin NX compute module supports fast onboard inference and is exploited for real-time inference of neural-networks.

\begin{figure*}
    \centering
    \newcommand{\imageheight}{100pt}
    \subfloat[\label{fig:autonomy_module:handheld}]{\includegraphics[height=\imageheight]{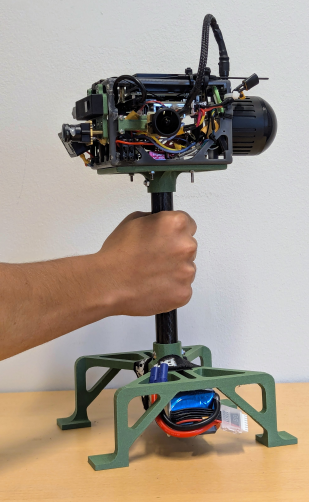}}
    \hfill
    \subfloat[\label{fig:autonomy_module:aerial}]{\includegraphics[height=\imageheight]{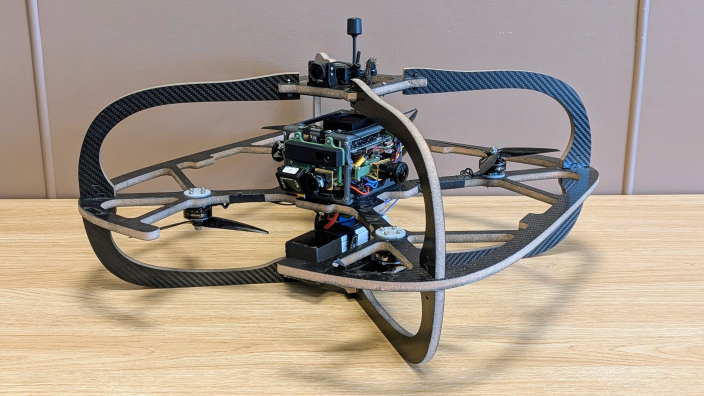}}
    \hfill
    \subfloat[\label{fig:autonomy_module:anymal}]{\includegraphics[height=\imageheight]{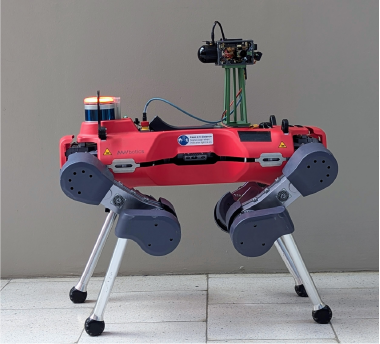}}
    \hfill
    \subfloat[\label{fig:autonomy_module:fixed_wing}]{\includegraphics[height=\imageheight]{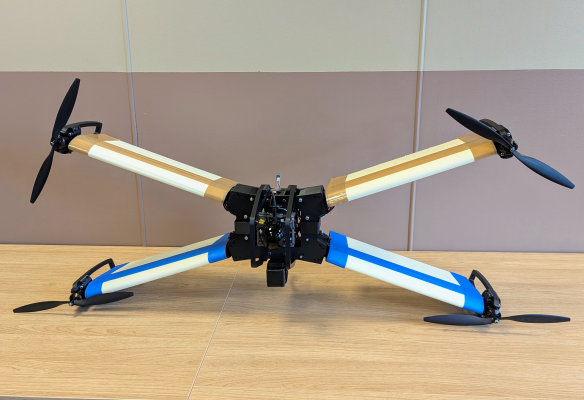}}
    \caption{The \moduleName{} autonomy module mounted on (a) handheld, (b) aerial, (c) legged, and (d) hybrid fixed-wing platforms.}
    \label{fig:autonomy_module}
\end{figure*}


\subsection{Diverse platforms}
\moduleName{} is designed to be a plug-and-play autonomy payload that can be integrated on diverse robot platforms. Through standardized UART or network interfaces, \moduleName{} can be interfaced with a diverse set of off-the-shelf control units and robotic platforms. The module is designed to be used as a standalone autonomy payload providing the solution for \ac{slam}, exploration and inspection path planning, neural-network based safety, and broadly capabilities for GPS-denied navigation in a single unit. At the same time it can function as a sensing and computing payload to run custom autonomy software.

We demonstrate the autonomy capabilities of \moduleName{} on both the custom built collision-tolerant aerial robot (\cref{fig:autonomy_module:aerial}) and the legged robot (\cref{fig:autonomy_module:anymal}. The frame of the collision-tolerant aerial robot is designed from a carbon-foam sandwich material. The robot has dimensions $\text{length} \times \text{width} \times \text{height} = \SI{0.52}{\meter} \times \SI{0.52}{\meter} \times \SI{0.24}{\meter}$, weighing \SI{1.472}{\kilogram} without the payload. \moduleName{} is interfaced with the mRo Pixracer Pro flight controller running PX4 firmware~\cite{meier2015px4}. Pose estimates and high-level commands such as position or velocity setpoints are relayed and tracked by the flight controller. The \moduleName{} module is also mounted on an ANYbotics ANYmal robot using a custom 3D-printed mount (\cref{fig:autonomy_module:anymal}) and interfaced with its onboard computers using Gigabit Ethernet. Position setpoints are provided to the ANYmal robot platform using \ac{ros}. The design of \moduleName{} also enables its use as a handheld data collection unit (\cref{fig:autonomy_module:handheld}). In addition to this, we present a the operation of the unit on a hybrid VTOL aircraft (\cref{fig:autonomy_module:fixed_wing}).

\section{EVALUATION}\label{sec:eval}
A set of experiments are conducted to demonstrate the enabling capabilities of the \moduleName{} across a variety of environments. We first evaluate the performance of the onboard~\ac{slam} solution in indoor environments for low-speed missions on the legged platform and the hybrid VTOL robot. Next, the performance of the method for high-speed flights is evaluated, followed by a comparison of the performance of the \ac{slam} solution with and without the radar sensor. The performance of the learning-based safe navigation policy is evaluated. Finally, exploration and inspection missions are performed using both the collision-tolerant aerial robot and the legged robot in a diverse set of environments.



\subsection{Robust \acf{slam}}
\begin{figure}[h]
    \centering
    \includegraphics[width=0.99\columnwidth]{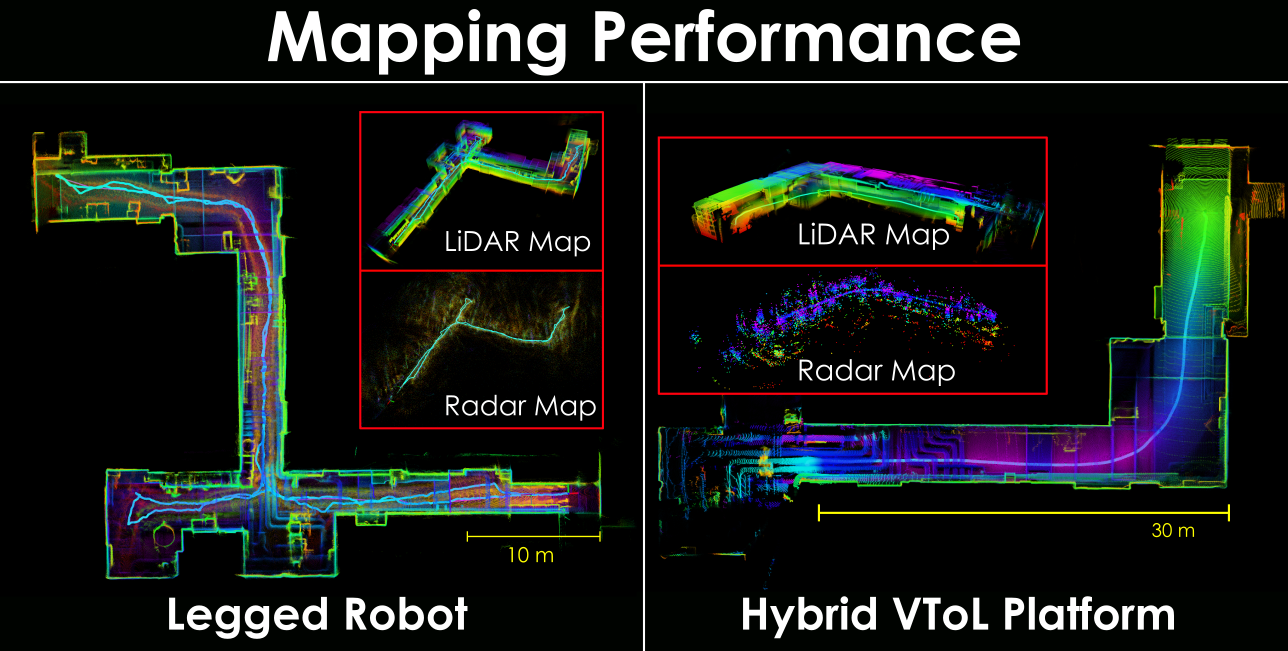}
    \vspace{-2ex}
    \caption{Mapping results with the legged and the hybrid VTOL platforms. }
    \label{fig:evaluation:diverse_robots}
\end{figure}

The mapping performance of the proposed unit is evaluated on the legged system and a hybrid VTOL platform. The hybrid VTOL is operated in a quadrotor mode. The maps of the environment and the aggregated radar point clouds are checked for consistency with the environment. The maps are shown in~\cref{fig:evaluation:diverse_robots}.

\begin{figure}[h]
    \centering
    \includegraphics[width=\linewidth]{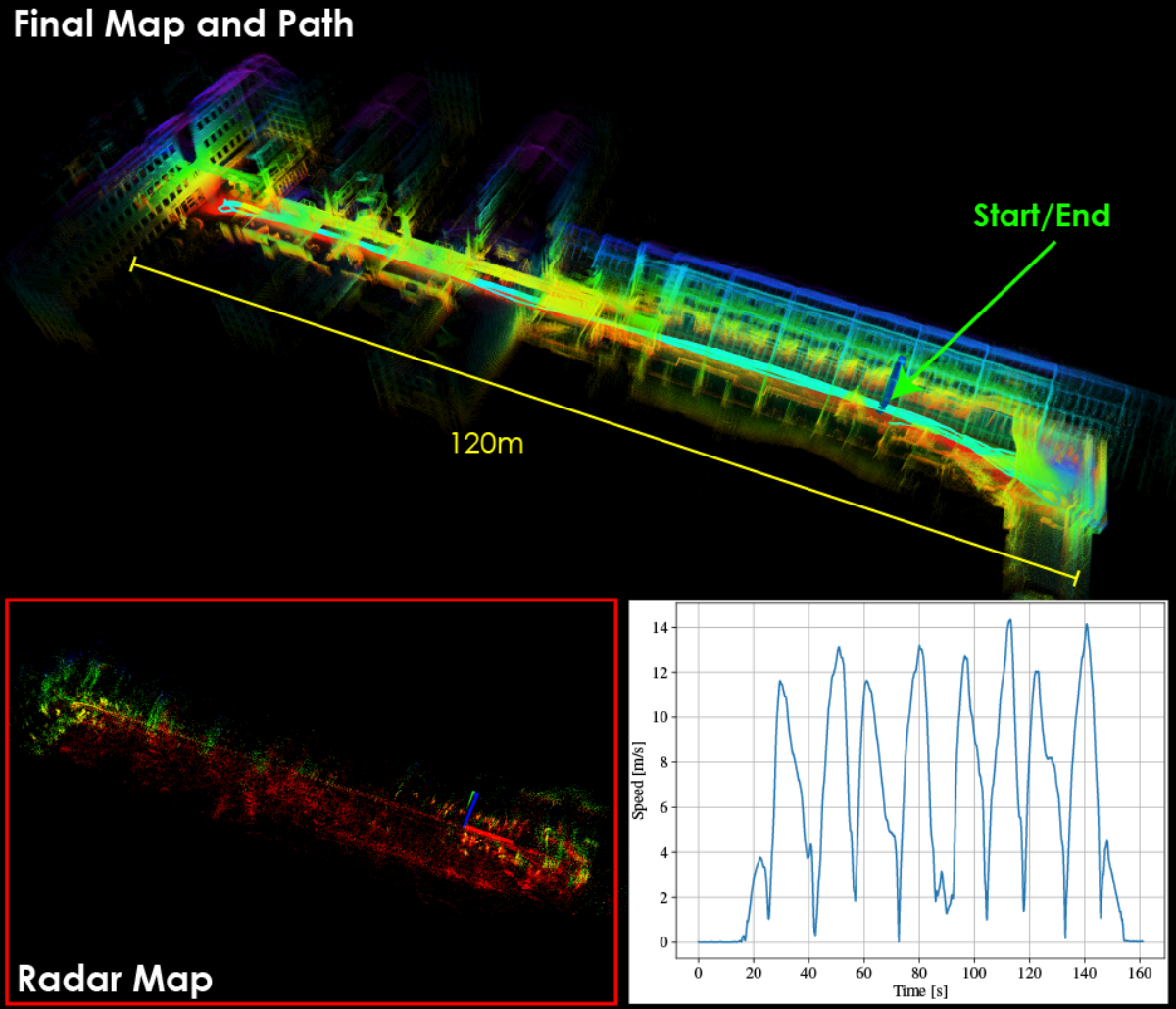}
    \vspace{-2ex}
    \caption{LiDAR map created online from the fast flight experiment (\qty{> 14}{\meter\per\second} according to the timeseries) of the aerial platform in the \ac{ntnu} corridor environment.}
    \label{fig:evaluation:fast_flight}
    \vspace{-3ex}
\end{figure}

Another experiment is conducted to evaluate the estimation performance during high-speed flights. The platform is manually piloted through an open-area of NTNU in Trondheim, Norway. In \cref{fig:evaluation:fast_flight}, the resulting map can be seen in this environment. The onboard estimate of the speed as a function of time can be seen in \cref{fig:evaluation:fast_flight}, which peaks at \qty{>14}{\meter\per\second}. The thin features of the windows and vertical beams demonstrate the accuracy of the \ac{slam} framework even at high speeds.


\subsection{Multi-modal estimation}
\begin{figure}[h!]
    \centering
    \includegraphics[width=0.99\columnwidth]{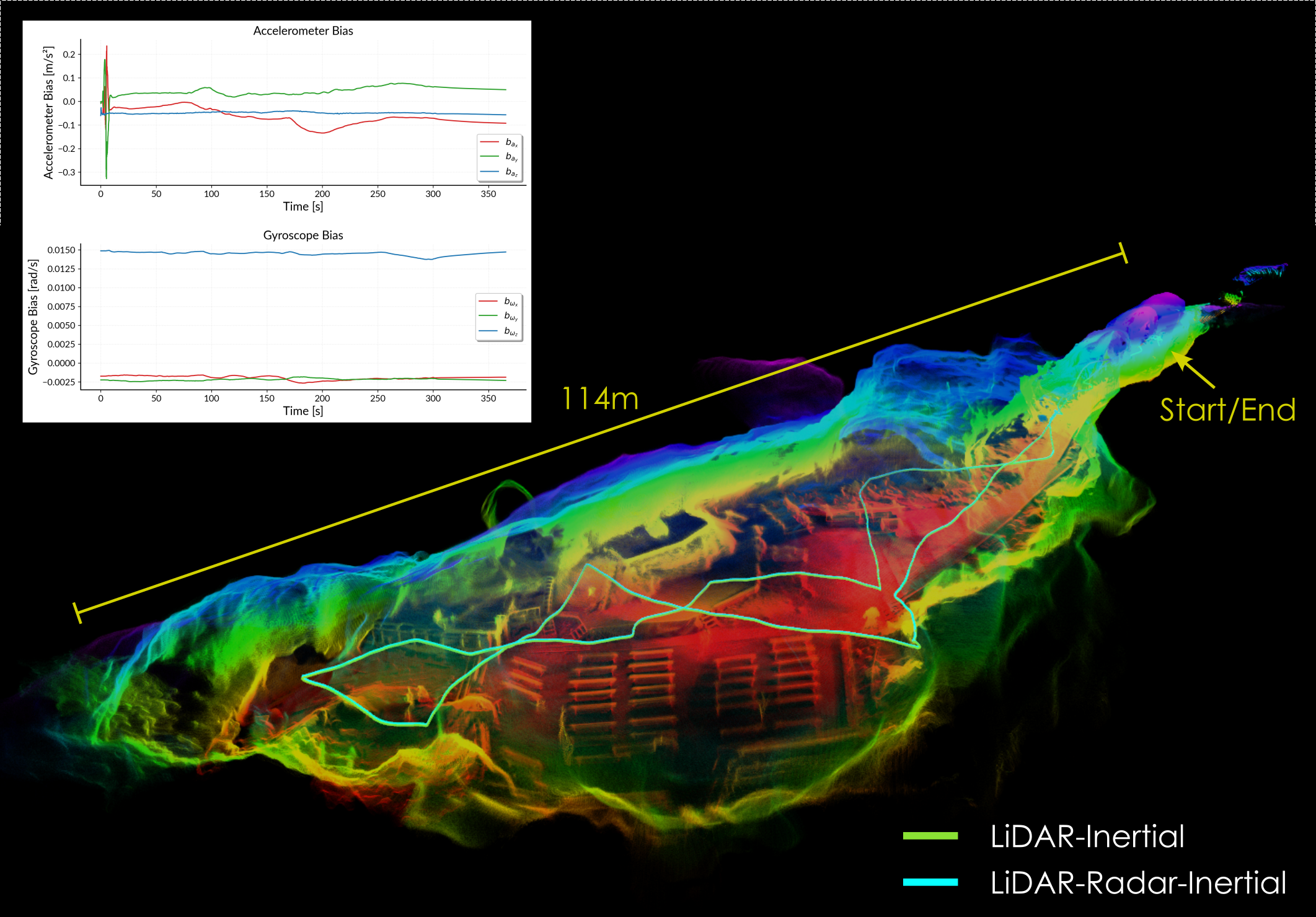}
    \caption{Accumulated registered LiDAR point clouds using the LiDAR-Radar-Inertial (Inset: estimated biases of the \ac{imu}) fusion and comparison of trajectories from LiDAR-Inertial and LiDAR-Radar-Inertial estimators.}
    \label{fig:evaluation:lidar_radar}
\end{figure}

In this experiment, we compare the result of fusing radar into the LiDAR-Inertial system as in~\cite{nissov2024degradation}. The resulting map from the LiDAR point clouds of the LiDAR-Radar-Inertial system are shown in Fig.~\ref{fig:evaluation:lidar_radar}. The trajectory estimated by the LiDAR-Radar-Inertial system is almost identical to that estimated by the LiDAR-Inertial method, with an absolute translational error of \qty{0.062 \pm 0.033}{\meter}. However, the addition of radar measurements has the additional benefit of adding resilience against visual obscurants and geometric self similarity as shown in~\cite{nissov2024degradation}.



\subsection{Neural Safety}
\begin{figure}[h!]
    \centering
    \includegraphics[width=\linewidth]{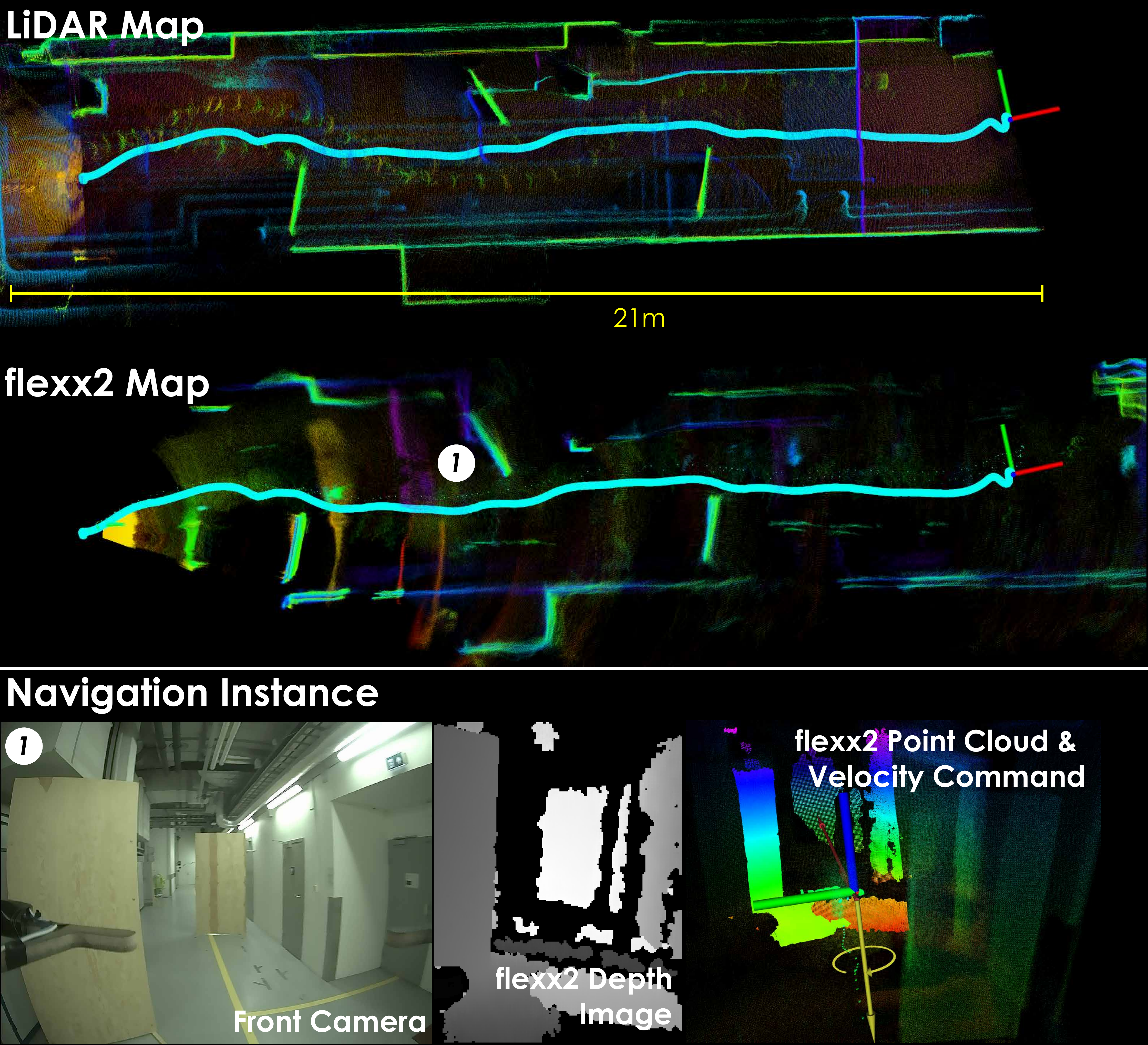}
    \caption{Overview of the neural safety experiment, showing an instance of the commanded velocity to avoid collision. Point clouds using data from flexx2 \ac{tof} and LiDAR sensors are shown, highlighting the disparity with respect to map quality and density.}
    \label{fig:evaluation:safety}
    \vspace{-2ex}
\end{figure}

This experiment showcases the neural safety policy deployed on the \moduleName{} module, which is mounted on the aerial platform from \cref{fig:autonomy_module:aerial}. The policy offers a layer of safety to the robot platform using purely raw sensor inputs from the \ac{tof} sensor and commanding velocity setpoints to the low-level flight controller. A simulation model of the robot with a \ac{tof} sensor is created in the Aerial Gym Simulator~\cite{kulkarni2025aerialgym} for parallelized simulation. A policy based on~\cite{kulkarni2024reinforcement} is trained in simulated room-like environments with floating objects. The network is tasked to reach a goal location across the room, navigating through dense clutter. The curriculum-based approach for environment population is used increase the clutter in the environment as the policy achieves desired success rates. The trained policy is transferred to \moduleName{} and raw data from the \ac{tof} sensor is used. Given a goal location, the policy commands velocity setpoints to the low-level flight controller to safely navigate the environment. This approach does not rely on access to prior maps, and is therefore not affected by accumulated drifts or mapping inconsistencies. The results from this experiment are presented in~\cref{fig:evaluation:safety}.

\begin{figure*}[t!]
    \centering
    \includegraphics[width=\linewidth]{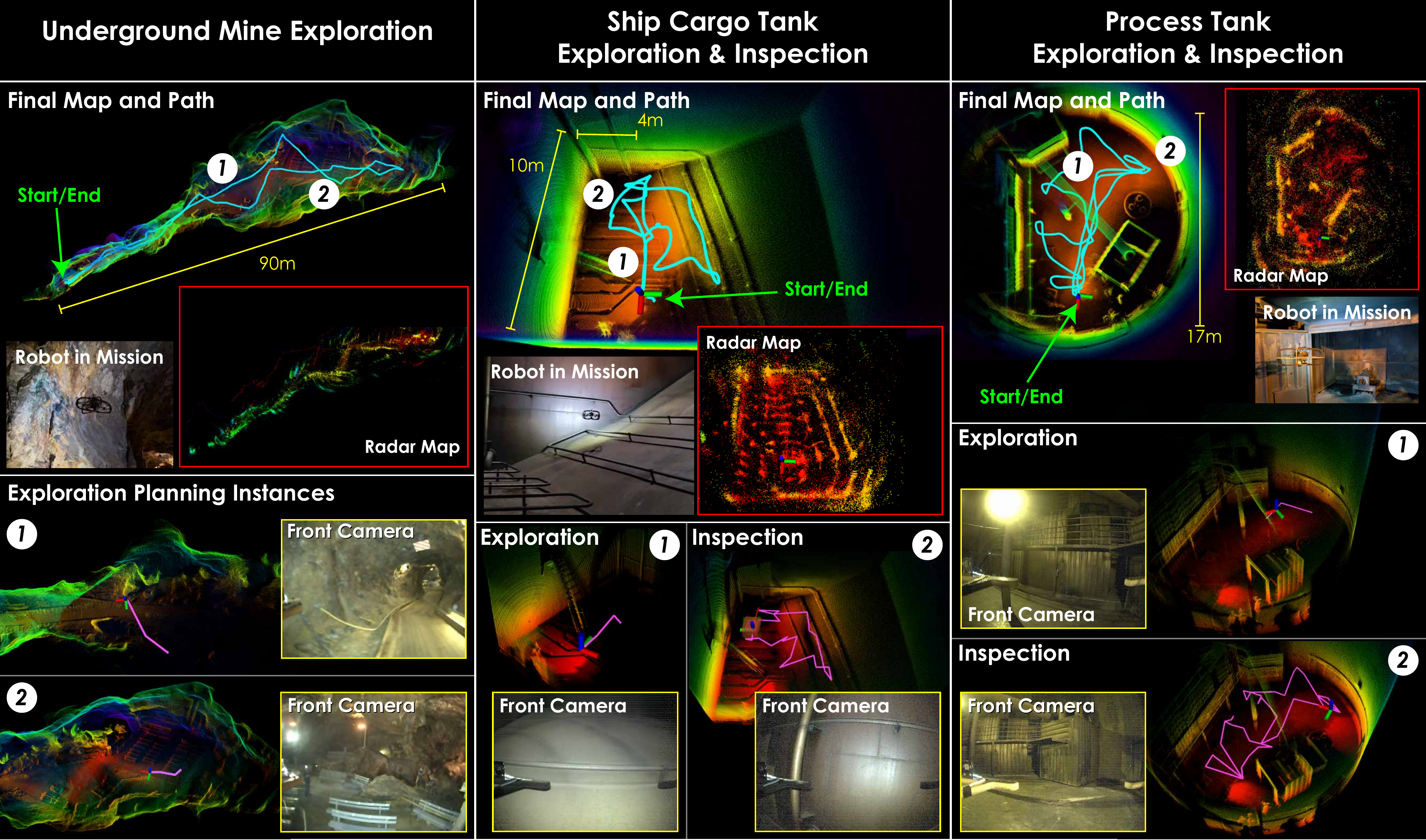}
    \caption{An overview of the autonomous exploration and inspection missions conducted in the L\o{}kken underground mine, cargo tank in an oil tanker vessel, and industrial process tank environments.}
    \label{fig:evaluation:exploration_inspection}
    \vspace{-3ex}
\end{figure*}
\subsection{Autonomous Exploration and Inspection}
This section presents the field deployments of the aerial and the legged platform~\cref{fig:autonomy_module:aerial} carrying \moduleName{} in diverse environments, conducting autonomous exploration and inspection missions, relying purely on onboard estimation with no prior map or any assumed knowledge of the environment.

\subsubsection{L\o{}kken Mine}
In this experiment, the robot was tasked to explore a section of the L\o{}kken underground mine in Tr\o{}ndelag, Norway. The robot started in a narrow tunnel section of the mine, entered the open area, and returned to the start location upon successfully completing the exploration. The exploration planner presented in ~\cref{sec:high_level_planning} was used. The mine presented a low-light environment with changing topology. \cref{fig:evaluation:exploration_inspection} (Left) presents the results from this test. A few exploration instances showing the paths and the onboard camera, the LiDAR and radar maps, and an instance of the robot are presented.

\subsubsection{Ship Cargo Hold}
In this experiment, the aerial platform from \cref{fig:autonomy_module:aerial} was tasked to conduct the volumetric exploration and visual inspection of a cargo hold inside an oil tanker ship. The tank was completely dark, and the robot relied purely on the onboard lights. The robot had no prior knowledge about the environment and was only given the rough bounds of exploration. The robot started in the \ac{ve} mode, as described in~\cref{sec:high_level_planning}, and conducted the exploration of the tank. Upon completion, it switched to the \ac{gvi} mode and calculated a path to inspect the surfaces of the mapped tank. The maximum height was restricted to $\SI{3}{\meter}$ for safety reasons. \cref{fig:evaluation:exploration_inspection} (Center) shows the results of this mission, including the LiDAR, radar maps, the robot in the environment, one instance of exploration, and the inspection.

\subsubsection{RelyOn}
In this experiment, the robot conducted the autonomous exploration and inspection of a process tank inside the RelyOn Nutec facility in Trondheim, Norway. The tank represents an industrial setting, with low-light conditions and low-visual-texture walls. Similar to the experiment in the cargo hold, the robot started in the \ac{ve} mode, explored the tank, and then switched to the \ac{gvi} mode for visual inspection. It is noted that the maximum height was restricted to \
\SI{3}{\meter} for safety reasons. The results from this mission are presented in \cref{fig:evaluation:exploration_inspection} (Right).

\begin{figure}[h!]
    \centering
    \includegraphics[width=0.99\columnwidth]{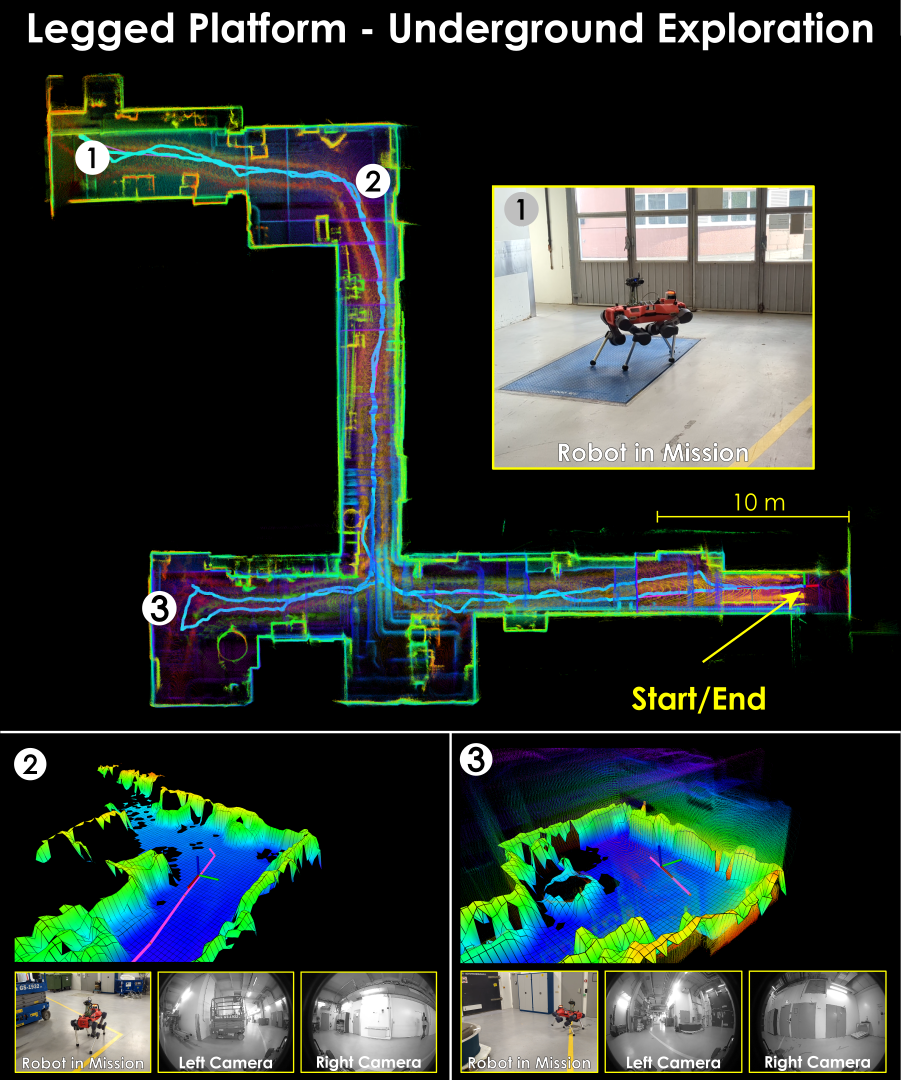}
    \caption{Autonomous exploration in an underground environment with the legged robot.}
    \label{fig:evaluation:anymalexploration}
\end{figure}

\subsubsection{Legged Robot}
This experiment presents the legged robot conducted an exploration mission of an underground environment at NTNU, Trondheim. The \moduleName{} unit is mounted using a raised platform on the robot (\cref{fig:autonomy_module:anymal}) to ensure that the LiDAR can see the ground around the robot. An elevation map is built using LiDAR measurements onboard the module and used to prune unsafe graph edges based on the work in~\cite{gbplanner3codebase} to ensure that the robot is commanded traversable paths for exploration. The robot autonomously explores the entire environment and returns to the start location as shown in~\cref{fig:evaluation:anymalexploration}.


\subsection{Synchronization}\label{sec:synchronization}
A bench-top experiment was conducted to analyze the consistency of the utilized time synchronization by comparing the theoretical sampling rates with the distribution of the actual sampling rates for each of the onboard sensors. An exception is made for the LiDAR, as the Airy does not use a constant sampling rate for the LiDAR point cloud, instead the LiDAR \ac{imu} timestamp results are provided to serve as a reference for the LiDAR point cloud timestamp accuracy. The experiment consists of capturing the sensor measurements for a duration of \qty{520}{\second} and then post processing them to analyze the distribution of sensor sampling rates. The results, noted in~\cref{tab:evaluation:synchronization}, show that the actual sampling rates have low-single-digit \unit{\milli\second} errors with respect to their corresponding theoretical values and a low standard deviation.

\begin{table}[h!]
    \centering
    \caption{Sensor Sampling Rates}
    \label{tab:evaluation:synchronization}
    \vspace{-2ex}
    \sisetup{
    table-format=3.3
}
\begin{tabular}{llSS}
    \toprule
    &\multirow{2}{*}{Sensor} &\multicolumn{2}{c}{Sample Time [\unit{\milli\second}]}\\
    &    &{Mean}	&{Std. Dev.}\\
    \midrule
    \multirow{3}{*}{\rotatebox{90}{\textbf{Misc.}}}

    &IMU	&5.101	&0.654\\
    &Radar	&100.000	&0.986\\
    &LiDAR IMU	&5.000	&0.005\\
    \midrule
    \multirow{4}{*}{\rotatebox{90}{\textbf{Cameras}}}
    &Front	&50.000	&1.286\\
    &Left	&50.000	&1.079\\
    &Right	&50.000	&1.080\\
    &ToF	&100.242	&0.880\\
    \bottomrule
\end{tabular}
\end{table}

\vspace{-2ex}

\section{CONCLUSIONS}\label{sec:conclusion}
This manuscript presented the design for \moduleName{}, a hardware-software module design to enable autonomy on robot embodiments across environments with varying degrees of degradation through multi-modal sensing. The hardware design decisions were outlined and the autonomy software utilized by the platform was described. An extensive set of experiments were conducted to evaluate the module mounted on aerial platforms, including high-speed flight, neural safety, as well as autonomous exploration and inspection. Autonomous exploration missions with the legged platform demonstrate the easy transferability and plug-and-play functionality of the module across platforms. The autonomous missions were conducted in different field environments, showcasing the capabilities of the platform in real scenarios.








\bibliographystyle{IEEEtran}
\bibliography{IEEEabrv,references}


\end{document}